\documentclass[letter, 10pt, conference]{ieeeconf}   

\IEEEoverridecommandlockouts
\usepackage{graphicx} 
                           
\usepackage{color}
\usepackage{mathtools}
\usepackage{siunitx}
\usepackage[backend=biber, style=ieee,maxbibnames=99]{biblatex}
\AtBeginBibliography{\footnotesize}
\newcommand{\edits}[1]{\textcolor{black}{#1}}

\title{\LARGE \bf
Stretchable Pneumatic Sleeve for Adaptable, \\Low-Displacement Anchoring in Exosuits}

\author{Katalin Schäffer$^{1,2}$, Ultan Fallon$^{3}$, and Margaret M. Coad$^{1}$%
\thanks{*This work was supported by the Ministry of Culture and Innovation of Hungary from the National Research, Development and Innovation Fund, financed under the TKP2021-NKTA funding scheme (project no. TKP2021-NKTA-66).}%
\thanks{$^{1}$Katalin Schäffer and Margaret M. Coad are with the Department of Aerospace and Mechanical Engineering, University of Notre Dame, Notre Dame, IN 46556, USA. 
{\tt\small \{kschaff2, mcoad\}@nd.edu}}%
\thanks{$^{2}$Katalin Schäffer is also with the Faculty of Information Technology and Bionics, Pázmány Péter Catholic University, 1083 Budapest, Hungary.}%
\thanks{$^{3}$Ultan Fallon is with the Department of Biomedical Engineering, University of Galway, Galway, Ireland. {\tt\small u.fallon2@nuigalway.ie}}%
}

\begin{document}

\maketitle
\thispagestyle{empty}
\pagestyle{empty}

\begin{abstract}
Despite recent advances in wearable technology, interfacing movement assistance devices with the human body remains challenging. We present a stretchable pneumatic sleeve that can anchor an exosuit actuator to the human arm with a low displacement of the actuator's mounting point relative to the body during operation. Our sleeve has the potential to serve as an adaptable attachment mechanism for exosuits, since it can adjust its pressure to only compress the arm as much as needed to transmit the applied exosuit forces without a large displacement. We discuss the design of our sleeve, which is made of fabric pneumatic artificial muscle (fPAM) actuators formed into bands. We quantify the performance of nine fPAM bands of various lengths and widths, as well as three sleeves (an fPAM sleeve, a series pouch motor (SPM) sleeve as in previous literature, and an off the shelf hook and loop sleeve), through the measurement of the compressing force as a function of pressure and the localized pulling force that can be resisted as a function of both pressure and mounting point displacement. Our experimental results show that fPAM bands with smaller resting length and/or larger resting width produce higher forces. 
Also, when inflated, an fPAM sleeve that has  equivalent dimensions to the SPM sleeve while fully stretched has similar performance to the SPM sleeve. 
While inflated, both pneumatic sleeves decrease the mounting point displacement compared to the hook and loop sleeve. Compared to the SPM sleeve, the fPAM sleeve is able to hold larger internal pressure before bursting, increasing its possible force range. Also, when not inflated, the fPAM sleeve resists the pulling force well, indicating its ability to provide anchoring when not actuated.
\end{abstract}

\section{Introduction}

With recent advances in soft robotics, the number of new designs for wearable devices has rapidly increased. In particular, a large variety of exosuit designs has been developed using various mechanisms and actuators to assist the movement of both the upper and lower limbs~\cite{Thalman2020_soft_wearable_review, Xiloyannis2021_soft_exosuit_review}. Interfacing these devices with the human body, however, remains challenging. Most often, hook and loop-based solutions are used, but there is much room for improvement in the ability of these designs to achieve a combination of user comfort, low displacement of actuator mounting points relative to the human body, and adaptability to the anchoring needs of various tasks. Recently, there has been interest in researching how methods of anchoring exosuits to the human body can be improved. For example, researchers have developed a cable-driven corset design that can actively vary its compressing force to only squeeze a user's arm when necessary and relax otherwise to avoid discomfort due to long-term pressure~\cite{Choi2019_active_anchor}. 

\begin{figure}[tb]
      \centering
      \includegraphics[width=8 cm]{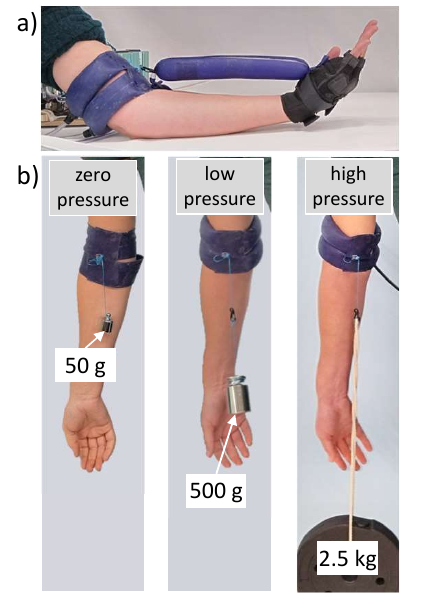}
      \caption{Our stretchable pneumatic sleeve anchored to the upper limb. (a) When integrated in an exosuit designed to move the human wrist, our pneumatic sleeve provides low-displacement anchoring for the exosuit's linear actuator on the upper arm. (b) Our pneumatic sleeve can adapt its compressing force by changing its internal pressure to only squeeze as hard as is needed to resist a desired pulling force.}
      \label{glamor_figure}
      \vspace{-0.5 cm}
   \end{figure}

Fluidically actuated structures show promise for use in anchoring exosuits to the human body, due to their ability to adapt to the shape of various users and distribute forces over large contact areas, which likely increases comfort. Various pneumatic or hydraulic structures have been used in haptic devices for physical human-robot interaction \cite{Do2021_pneumatic_pouches_wearable_haptic, Nunez2022_large_area_haptic_device, Young2019_Bellowband}, as well as for other applications such as compression therapy~\cite{Zhu2023_peristaltic_compression_therapy}. Recently, a pneumatic sleeve for upper limb exosuit anchoring was developed that can actively change its diameter and compressing force by varying its internal pressure~\cite{Diteesawat2022_pneumatic_band}. This sleeve design is based on forming a series pouch motor (SPM) linear contractile actuator~\cite{Niiyama2014_SPM, coad2019vine} into a band that wraps around the user's arm. While this design is promising for exosuit anchoring, there is still much knowledge to be gained by exploring the performance of additional anchoring methods.

In this paper, we present a novel stretchable pneumatic sleeve for adaptable, low-displacement exosuit anchoring on the upper limb (Fig.~\ref{glamor_figure}). Our sleeve is based on forming fabric pneumatic artificial muscle (fPAM) linear contractile actuators~\cite{Naclerio2020_fPAM} into bands that wrap around the user's arm. The fPAM is made of an airtight ripstop nylon fabric which has a regular grid weave pattern, therefore it stretches along a 45$^\circ$ angle to its weave but not directly along the fibers. In contrast to an SPM, which is made of inextensible fabric that is partially sealed along its length and contracts when inflated due to the various pouches along its body ballooning up and shortening, an fPAM is formed into a simple tube shape (with the weave at 45$^\circ$), and it contracts along its length and extends radially upon inflation due to the directional extensibility of its material. The fPAM is similar to a McKibben artificial muscle~\cite{gaylord1958fluid}, but because it consists of a single layer of fabric, it is foldable and low-hysteresis, and it shows a quick dynamic response.
\edits{Although fPAMs have been used in various soft robot designs~\cite{selvaggio2020obstacle,fPAM_exo}}, it is unknown how fPAM actuators behave in a band form, and how an fPAM-based sleeve compares in performance to an SPM-based sleeve or a more typical hook and loop anchoring method, so that is the focus of this work.

In the rest of this paper, we first introduce the design and fabrication of our fPAM bands and the sleeves used in this work, including the parameters used to describe the designs. We then discuss our experimental setup and measurement process for evaluating the performance of various designs and present the results of several experiments. First, we compare the performance of fPAM band designs with various initial lengths and widths. Next, we compare the performance of an SPM band with that of two similar fPAM bands. Then, we compare the performance of three sleeves: one made from fPAM bands, one made from an SPM band as in~\cite{Diteesawat2022_pneumatic_band}, and one off the shelf hook and loop-based sleeve as in~\cite{fPAM_exo}. Finally, to explore the effect of the various sleeves on exosuit performance, we present a demonstration of all three sleeves integrated into our soft exosuit that uses an fPAM as a linear contractile actuator to move the human wrist~\cite{fPAM_exo}. In this application, it is especially important to reduce the mounting point displacement to allow the full force of the linear fPAM to be used for actuation.



\section{Design and Fabrication}

\subsection{Parameters of the fPAM Band}
\begin{figure}[tb]
      \centering
      \includegraphics[width=\columnwidth]{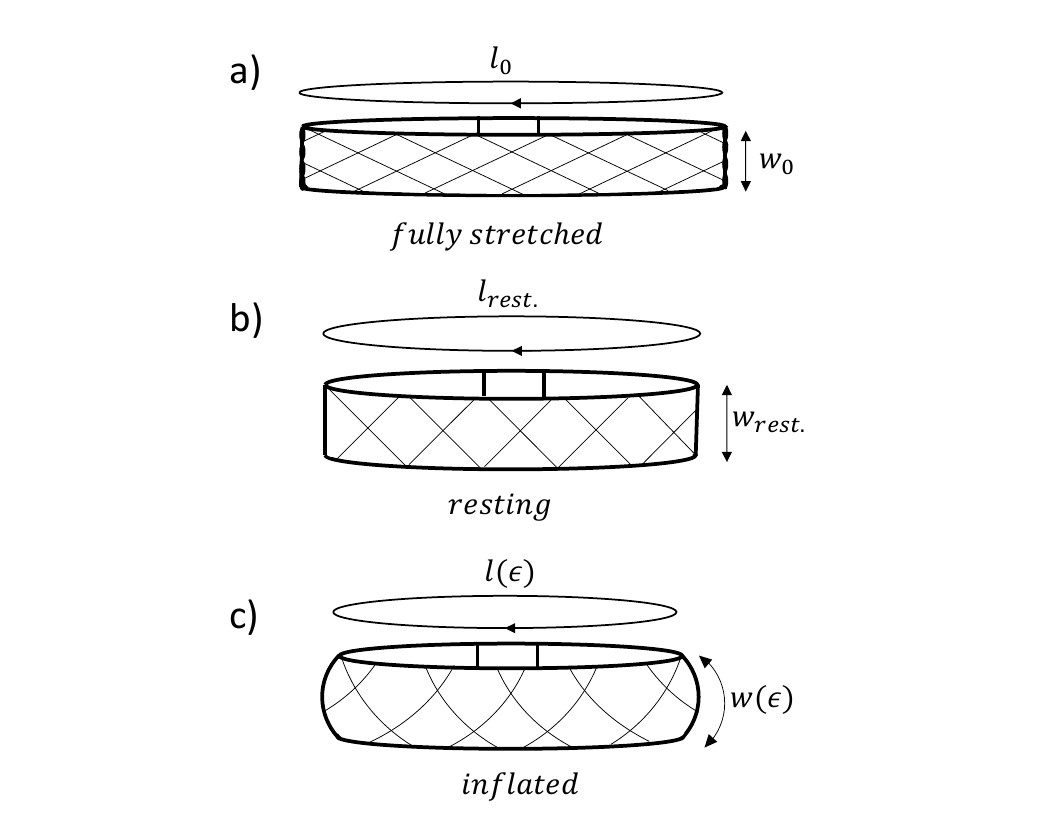}
\vspace{-0.6 cm}      \caption{Schematic of the states of the fabric pneumatic artificial muscle (fPAM) band that makes up our sleeve. (a) In the fully stretched state, the fPAM band is uninflated, and the band fabric is stretched to reach its maximum length $l_0$ and minimum width $w_0$. 
(b) In the resting state, the fPAM band is uninflated and unstretched, with length $l_{rest.}$ and width $w_{rest.}$. These resting parameters are used to describe the width and length of the fabricated bands. (c) In the inflated state, the fPAM band is inflated, and its length $l(\epsilon)$ and width $w(\epsilon)$) are functions of the contraction ratio $\epsilon$.}
      \label{geometric_model}
      \vspace{-0.5 cm}
   \end{figure}

Here, we describe the geometry of the fPAM band (fully stretched and resting state) and its shape change when pressurized. Fig.~\ref{geometric_model} shows the schematic drawing of the sleeve when deflated and stretched, when deflated and resting, and when inflated, along with the corresponding notations. When in a linear contractile actuator form as first presented in~\cite{Naclerio2020_fPAM}, fPAM actuators are at their longest when at their initial, fully stretched length, $l_0$. At this length, their width $w_0$ is at its smallest. At rest, fPAM actuators have a shorter length $l_{rest.}$ and a wider width $w_{rest.}$ than their initial length and width. When inflated, the length of the fPAM $l(\epsilon)$ shortens and the width $w(\epsilon)$ increases until the maximum contraction ratio $\epsilon$ is reached.

\begin{table}[b]
\caption{Dimensions of the band and sleeve prototypes\label{tab:prototypes_parameters}}
\vspace{-0.2 cm}
\centering
\begin{tabular}{||p{2.4 cm} p{0.7 cm} p{0.8 cm} p{2.8 cm}||}  
     \hline
      name & width [cm] & length [cm] & configuration\\ 
     \hline\hline
     nine fPAM bands   & 3,4,5~~~$\times$ & 27,28,30 & 1 chamber, single tube\\ 
     \hline
     $\rightarrow$ fPAM(r) band  & 4 &   30 & 1 chamber, single tube\\
     \hline
     $\rightarrow$ fPAM(s) band  & 5 & 27 & 1 chamber, single tube\\
     \hline
     SPM band   & 4   & 30 & 1 chamber, 7 pockets\\ 
    \hline\hline
     SPM sleeve  & 8 &  32  & 1 chamber, 7 pockets\\
     \hline
     fPAM sleeve  & 2$\times$5 & 28 & 2 chambers, double tube\\
    \hline
\end{tabular}
\end{table}

\subsection{Pneumatic Bands}

\begin{figure}[tb]
      \centering
      \includegraphics[width=\columnwidth]{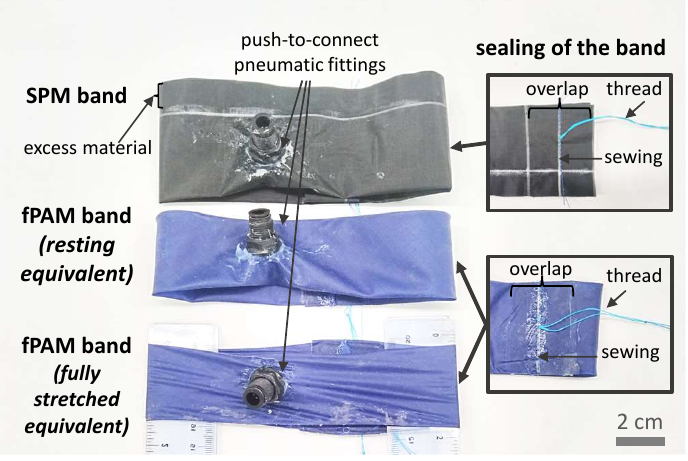}
      \caption{The series pouch motor (SPM) band \textit{(top)} and the two equivalent fPAM bands tested. The \edits{fPAM(r)} band \textit{(middle)} has the same length and width as the SPM band when resting, and the \edits{fPAM(s)} band \textit{(bottom)} has the same dimensions when it is fully stretched as shown. All the bands have a push-to-connect pneumatic fitting to allow inflation and have an overlapping area of 4~cm where the ends of the band are sewn together. In the middle of the overlapping area, a thread is attached for the pulling force and holding force measurement.}
      \label{fabricated_bands}
      \vspace{-0.5 cm}
   \end{figure}

We fabricated nine pneumatic fPAM bands with three resting widths (3~cm, 4~cm, and 5~cm) and three resting lengths (27~cm, 28~cm, and 30~cm) to examine how the change of the sleeve design influences the performance of the anchoring. Additionally, we made an SPM band to compare its performance with a fully stretched and resting equivalent fPAM band, as shown in Fig.~\ref{fabricated_bands}. The fabricated SPM band is 4~cm wide and 30~cm long, and it has the same dimensions as one of the fPAM bands in its resting state \edits{(denoted as fPAM(r))}. The fPAM band with 5~cm resting width and 27~cm resting length \edits{(denoted as fPAM(s))} has equivalent dimensions to the SPM band when it is fully stretched. \edits{The parameters of all the bands tested are listed in Table~\ref{tab:prototypes_parameters}.} 

The fPAM bands were made from silicone-coated ripstop nylon fabric (30 Denier Double Wall Ripstop Nylon Silicone Coated Both Sides, Rockywoods). \edits{A rectangle-shaped piece was cut out of the fabric with a 45$^\circ$ fiber orientation and then glued together (Sil-Poxy, Smooth-On) along its long side with approximately 1.5~cm overlap to form a tube as described in~\cite{Naclerio2020_fPAM}. The ends of the tube were sealed flat with the glue.} The SPM band was made from TPU-coated heat-sealable ripstop nylon fabric (40 denier, extremtextil, Dresden, Germany). \edits{Similarly to the fPAM, the SPM band was fabricated from a rectangular piece of fabric and, first, it was manually heat-sealed along its long sides with 2~cm excess material to form a tube. Then, seven square-shaped pouches were made by sealing approximately 2 cm long lines in the middle of the tube perpendicular to the seal on the side.}

All the bands have a push-to-connect pneumatic fitting and they have an overlapping area of 4 cm where the ends of the band are sewn together. In addition to the sewing, the fPAM bands are also glued together, which provides a non-stretchable region to which actuators can be attached. We sewed a thread of fishing line to the midpoint of the overlapping area, which serves as the mounting site on the bands when measuring the holding force.

\subsection{Three Sleeve Designs}

\begin{figure}[tb]
      \centering
      \includegraphics[width=\columnwidth]{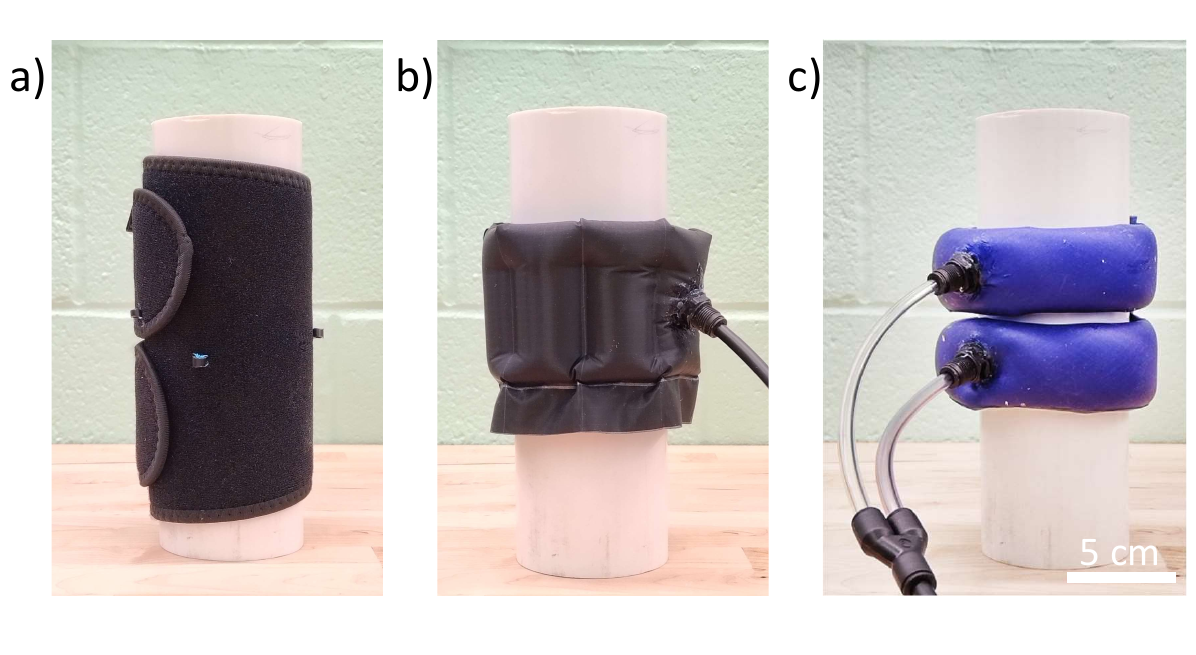}
      \vspace{-0.9 cm}
      \caption{The three sleeves tested, all placed on a cylinder of 7.3~cm diameter: (a) the off the shelf hook and loop sleeve, (b) the SPM sleeve, and (c) the fPAM-based sleeve designed to be equivalent to the SPM sleeve when fully stretched.}
      \label{three_sleeves}
      \vspace{-0.4 cm}
   \end{figure}

One of the sleeves that we examined is the commercially available hook and loop sleeve (Elbow Support Sleeve Brace, Bracoo) (Fig.~\ref{three_sleeves}(a)). This sleeve is made of an elastic fabric. When it is tightened around the arm at the elbow, it passively provides static anchoring. This sleeve was previously used in our exosuit design \cite{fPAM_exo} with small hoops sewn to it to serve as mounting points.

The other two sleeves are pneumatically actuated sleeves that can provide adaptive anchoring as their tightness is regulated by pressure. The first pneumatic sleeve is the SPM sleeve (Fig.~\ref{three_sleeves}(b)), which is based on the design introduced in~\cite{Diteesawat2022_pneumatic_band}. The sleeve was fabricated the same way as the SPM band, but with a width of 8~cm and length of 32~cm; it has seven pouches. The second pneumatic sleeve is made of two fPAM bands (Fig.~\ref{three_sleeves}(c)), each of which is equivalent in dimensions to the half-width SPM sleeve in their fully stretched state. Because we expect that the compressing force of the SPM band is proportional to its width, we used two parallel bands in the fPAM sleeve design instead of doubling the diameter the fPAM band. This also helps to maintain a low profile sleeve configuration. The resting width of the bands is 5~cm and the resting length is 28~cm. The two fPAM bands were attached together by gluing two extra layers of fabric to the two sides of the overlapping area of the bands. \edits{The parameters of the pneumatic sleeves are listed in Table~\ref{tab:prototypes_parameters}.}

\section{Experimental Procedure}
To quantify the performance of the fabricated bands and sleeves, we followed similar evaluation methods as described in~\cite{Diteesawat2022_pneumatic_band}. The two quantities that we measured for all prototypes were the compressing force \edits{and the holding force. Furthermore, we characterised the stiffness of the sleeves by measuring the pulling force as a function of the mounting point displacement}. The compressing force is the force that the band or sleeve applies radially to the arm; this quantifies how firm the attachment is. The holding force is the pulling force that we need to apply at the mounting point to reach 2~cm displacement along the arm. We specify this pulling direction as 0$^{\circ}$ compared to the centerline of the arm or test cylinder. The \edits{stiffness of the full sleeves} at the mounting site can be further analyzed by measuring the pulling force as a function of the displacement of the mounting point. We conducted this measurement on the full sleeves along pulling directions of 0$^{\circ}$ and 45$^{\circ}$, which corresponds to the approximate range of the angles between the actuator fPAM and the human arm in our previously developed wrist exosuit~\cite{fPAM_exo}.

\subsection{Experimental Setup}

\begin{figure}[tb]
      \centering
      \includegraphics[width=8.0 cm]{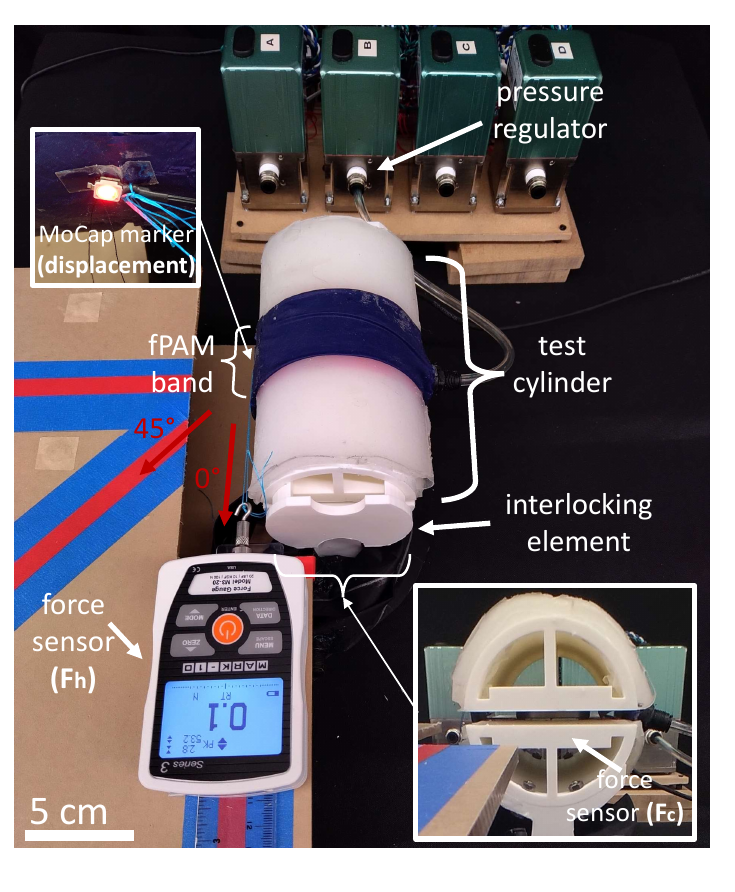}
      \vspace{-0.6 cm}
      \caption{Measurement setup for compressing force, holding force and mounting point displacement measurement. The setup consists of a test cylinder around which the tested band or sleeve is wrapped, as well as an elevated platform for positioning the external force sensor, which can measure the pulling force or holding force ($F_h$) at the mounting point. 
      A closed-loop pressure regulator maintains the desired pressure in the band or sleeve. The displacement of the mounting point is measured by a motion capture marker \textit{(top left corner)}, and the compressing force applied by the band ($F_c$) is measured by the force sensor in the center of the cylinder \textit{(bottom right corner)}. The interlocking elements at the ends of the cylinder can be added or removed depending on which force measurement is currently taking place.}
      \label{experimental_setup}
      \vspace{-0.5 cm}
   \end{figure}

Fig.~\ref{experimental_setup} shows the experimental setup with one of the fPAM bands. \edits{One of the pressure regulators from the pressure regulator board (top) was used to set the desired pressure in the band. The board also consists of a signal conditioning circuit and micro-controller (Arduino Uno) (not visible on the picture). We designed and 3D printed a cylinder (center) with 8~cm diameter and covered it with a 0.5~cm silicone layer (Ecoflex 00-30, Smooth-On) on its surface to create a proxy for the arm of a human user.} The setup also includes an elevated platform, which helps to position the force sensor (M3-20, MARK-10) when pulling on the mounting point of the band. For the \edits{stiffness characterization of the full} sleeves, the displacement of the mounting point was measured by placing a motion capture marker (PhaseSpace, Impulse X2E) on the sleeve, directly above the mounting point as shown in the top left corner of Fig.~\ref{experimental_setup}.

The test cylinder is made of two half-cylinders with a force sensor (Mini 45, ATI) placed between them at the center (bottom right corner of Fig.~\ref{experimental_setup}). The force sensor records the compressing force exerted by the band or sleeve on the cylinder. When other measurements were conducted, interlocking structures were used to lock the two parts of the cylinder together as shown in Fig.~\ref{experimental_setup}.

\subsection{Measurement Process}

 For the compressing force measurement, the force sensor within the two half cylinders was used (without the interlocking elements). The sensor was set to zero before the band or sleeve was placed on the test cylinder to avoid the offset from the weight of the top half cylinder. The band or sleeve was placed to be centered around the middle of the cylinder. For the pneumatic bands and sleeves, the compressing force was recorded at 0~kPa, 13.8~kPa, 27.6~kPa, and 41.1~kPa pressure, and we repeated each measurement three times.

For the holding force measurement, the cylinder was locked at the two ends by using the interlocking elements (Fig.~\ref{experimental_setup}). The band was placed in the middle of the cylinder, and we manually pulled on the thread attached to the mounting point of the band along the 0$^{\circ}$ direction on the platform using the force sensor. The initial position of the force sensor was set when we measured 0.1~N initial tension, and then we manually pulled the sensor \edits{along a ruler} until we reached 2~cm displacement from the initial position. We record the pulling force measurement from the sensor for the four previously described pressure levels and repeated each measurement three times.

The \edits{stiffness characterization} measurement for the full sleeves was conducted similarly to the holding force measurement for the bands but with the change of recording force data \edits{over time using MATLAB} and monitoring the position of the mounting point with the attached motion capture marker.  \edits{We characterized the stiffness by measuring} the pulling force along the pulling directions of 45$^{\circ}$ and 0$^{\circ}$ in relation to the mounting point displacement. The measurement was repeated for four different pressure levels for the fPAM sleeve and three different pressure levels for the SPM sleeve as it could not reach the highest pressure level. \edits{The measurements were synchronized by time stamping both sets of data. The motion capture data was time-stamped by taking a screenshot with the current time and recording time, and the pulling force data was time-stamped when recorded through MATLAB.} Also, we measured the compressing force that is applied to the arm when the sleeves operate at the examined pressure levels.


\section{Results and Discussion}

\subsection{Comparison of the fPAM Bands}

\begin{figure}[tb]
      \centering
      \includegraphics[width=\columnwidth]{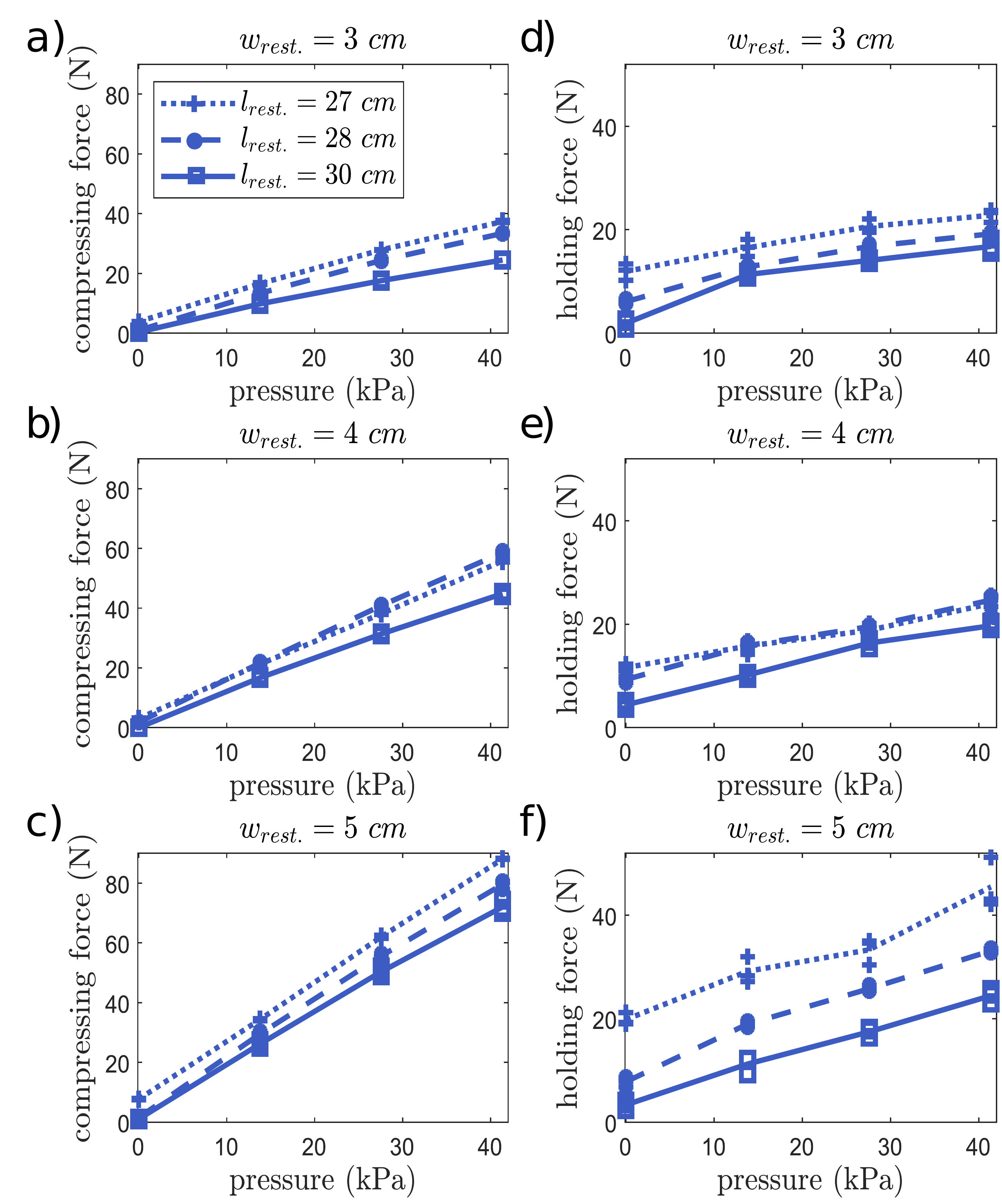}
      \vspace{-0.7 cm}
      \caption{Measurement results for the nine fPAM bands with three widths and three lengths. (a-c) Compressing force of the bands as a function of their internal pressure (organized into separate figures according to the band's width), and (d-f) holding force of the bands as a function of their internal pressure (organized into separate figures according to the band's width).}
      \label{fPAM_bands_measurement_results}
      \vspace{-0.5 cm}
   \end{figure}

 The results of the compressing force and holding force measurements on the nine fPAM bands are shown in Fig.~\ref{fPAM_bands_measurement_results}. The measured force is plotted as a function of the applied pressure, with the graphs organized into separate figures according to the band's width.

The compressing force measurements (Fig.~\ref{fPAM_bands_measurement_results} (a-c)) show that the increase of the band width and the decrease of the band length increase the magnitude of the compressing force. The effect of the width change on the force is expected, as the increased width increases the surface area of the band. The effect of the length change is also straightforward based on the force equation of the fPAM actuator, as presented in~\cite{Naclerio2020_fPAM}. The shorter fPAM band operates at a smaller contraction ratio (i.e., closer to its fully stretched length), where the fPAM force is higher. The measurements results, however also show an initial offset at zero pressure for the shortest bands of 27~cm length. The circumference of the test cylinder is 28.3~cm, therefore the shortest band is stretched when it is put on the test setup, and due to the elasticity of the material it applies compressing force when the band is not inflated.

The measurement results in Fig.~\ref{fPAM_bands_measurement_results} (d-f) show that the holding force scales similarly to the compressing force when the width and the length of the fPAM band changes. The benefit of the tight fit band becomes more prominent when the width of the band is increased. When the average magnitude of the holding force for the 3~cm width bands and for the 5~cm with bands is compared, we can observe a 28\% increase for the longest bands, a 56\% increase for the mid-length bands, and a 78\% increase for the short bands.

\subsection{Comparison of the SPM and fPAM Bands}

\begin{figure}[tb]
      \centering
      \includegraphics[width=\columnwidth]{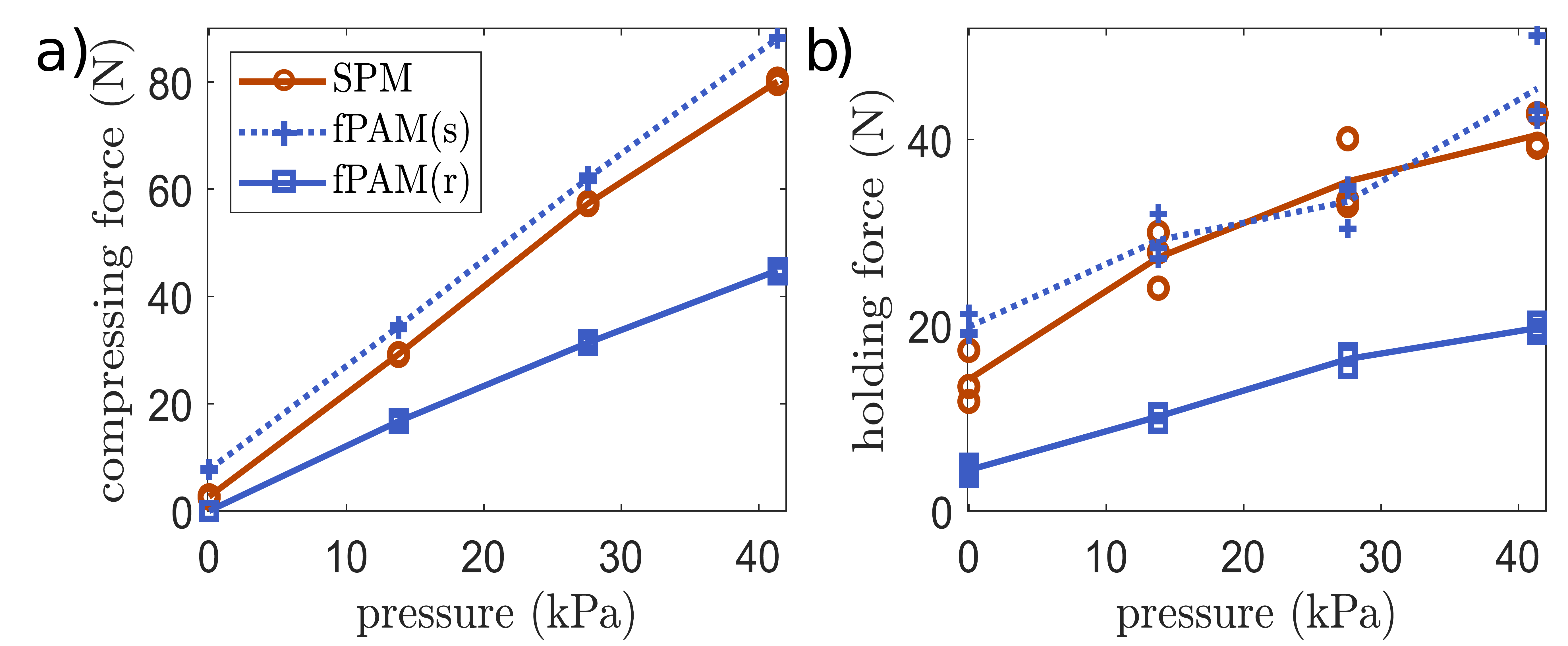}
      \vspace{-0.7 cm}
      \caption{Measurement results for the SPM band and the two equivalent fPAM bands, the fully stretched equivalent (fPAM(s)) and the resting equivalent (fPAM(r)), shown in Fig.~\ref{fabricated_bands}. (a) Compressing force of the bands as a function of their internal pressure, and (b) holding force of the bands as a function of their internal pressure.}
      \label{fPAM_vs_SPM_band}
      \vspace{-0.6 cm}
   \end{figure}

We conducted the same compressing force and holding force measurements on the SPM band. Fig.~\ref{fPAM_vs_SPM_band} shows the plot of the measurement results in comparison with the two equivalent fPAM bands that were presented in Fig.~\ref{fabricated_bands}.

The SPM band demonstrates comparable performance to the fully stretched equivalent fPAM band in both compressing and holding forces. However, the fully stretched equivalent fPAM band exhibits higher forces at zero pressure owing to its elasticity. Despite having the same resting length and width as the SPM band, the resting equivalent fPAM band generates lower compressing and holding forces compared to other bands. These measurement results show that the SPM band shares similar characteristics with the fPAM band when the fPAM band in its fully stretched state has equivalent dimensions to the SPM band.

\subsection{Comparison of the Three Sleeves}

\begin{figure}[tb]
      \centering
      \includegraphics[width=\columnwidth]{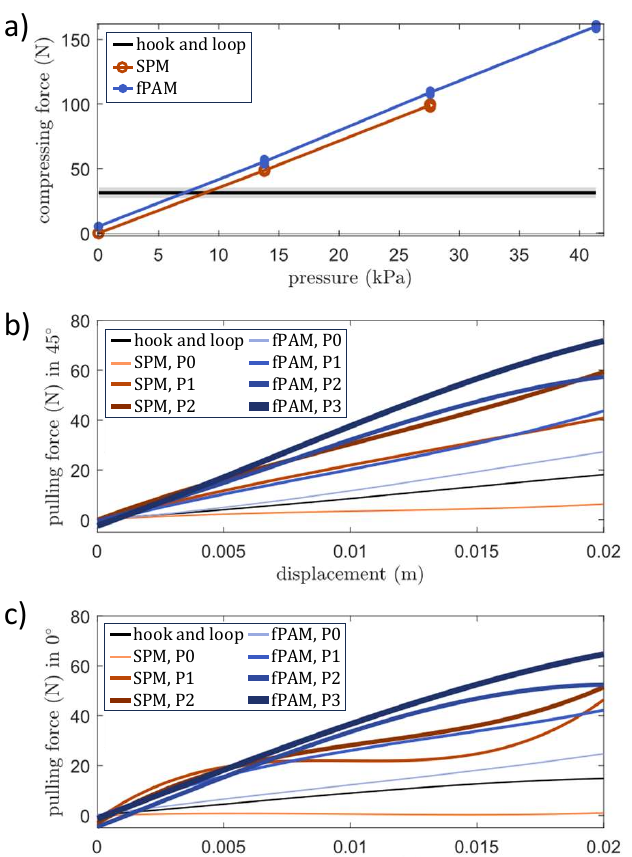}
      \caption{Measurement results for the three sleeves shown in Fig.~\ref{three_sleeves}. (a) Compressing force of the sleeves as a function of their internal pressure, (b) pulling force as a function of the mounting point displacement in the 45$^\circ$ direction for all the sleeves and at different pressure levels \edits{(P0~=~0~kPa, P1~=~13.8~kPa, P2~=~27.6~kPa and P3~=~41.4~kPa)}, and (c) pulling force as a function of the mounting point displacement in the 0$^\circ$ direction for all the sleeves and at different pressure levels.}
      \label{sleeve_measurement_results}
      \vspace{-0.5 cm}
   \end{figure}

We conducted a comparison of the three distinct sleeve designs (Fig.~\ref{three_sleeves}), with the SPM-based sleeve and \edits{the stretched} fPAM sleeve being dimensionally identical, and the results are shown in Fig.~\ref{sleeve_measurement_results}.

As shown in Fig.~\ref{sleeve_measurement_results}(a), the compressing force exerted by the fPAM sleeve closely matches that of the SPM sleeve, confirming that the fPAM sleeve can achieve similar compression levels under the same pressure. Additionally, the measurement results validate our hypothesis that the compressing force linearly scales with the width of the SPM sleeve and with the number of the parallel fPAM bands for the examined sleeves. The measurement results also display the fPAM sleeve's ability to exert some level of compressing force even at zero applied pressure. We observed that the fPAM  band can hold higher pressure than the SPM band which popped due to failure of the pouch heat seals when the pressure went over 27.6~kPa. Also, both pneumatic sleeves can provide higher compressing force \edits{than the hook and loop sleeve} when the applied pressure is higher than approximately 8 kPa.

As shown in Fig.~\ref{sleeve_measurement_results}(b) and (c), we observed a consistently higher magnitude of pulling force in the 45$^{\circ}$ direction compared to the 0$^{\circ}$ direction across all sleeves. This difference might be attributed to the reduced sliding when the pulling force deviates from 0$^{\circ}$.
In most cases, the force-displacement relationship follows a close-to-linear pattern, except for the SPM band at 0$^{\circ}$, where the force shows saturation. This should be further examined through a more in-depth analysis considering the material properties involved.

The force-displacement behavior of the fPAM band matches that of the SPM sleeve for nonzero pressures. For the fPAM sleeve, higher pressures can be applied, which reduces the mounting point displacement. However, defining a maximum pressure threshold is crucial to ensure user safety. Further research should explore the potential advantages offered by the extended capabilities of the fPAM-based pneumatic sleeve while maintaining user safety.


Both the hook and loop sleeve and the fPAM sleeve are made of elastic materials, ensuring a tight fit. The \edits{uninflated} fPAM sleeve experiences less displacement than the hook and loop sleeve, likely due to the mounting point's position within the overlapping area of the band, which possesses higher rigidity.
   
\section{Demonstration of the Three Sleeves on an fPAM-Actuated Exosuit}

\edits{In this section, we demonstrate the anchoring abilities of the three sleeves in a wrist exosuit with contractile actuation.}

\edits{The demonstrated exosuit consists of a single actuator fPAM attached to a glove on the hand and to the sleeve on the arm as shown in Figs.~\ref{demonstration1} and~\ref{demonstration2}}. We placed each sleeve on the upper arm of the user, close to the elbow. \edits{This placement allows us to better observe the sleeve deformation compared to placing it on the forearm, which is the typical placement for a wrist exosuit~\cite{fPAM_exo}. We placed motion capture markers on the arm to monitor the elbow and wrist angle, as well as one motion capture marker on the sleeve, next to the mounting point.} 

\subsection{Comparison of Three Sleeves with Fixed Wrist Position}

\begin{figure}[tb]
      \centering
      \includegraphics[width=\columnwidth]{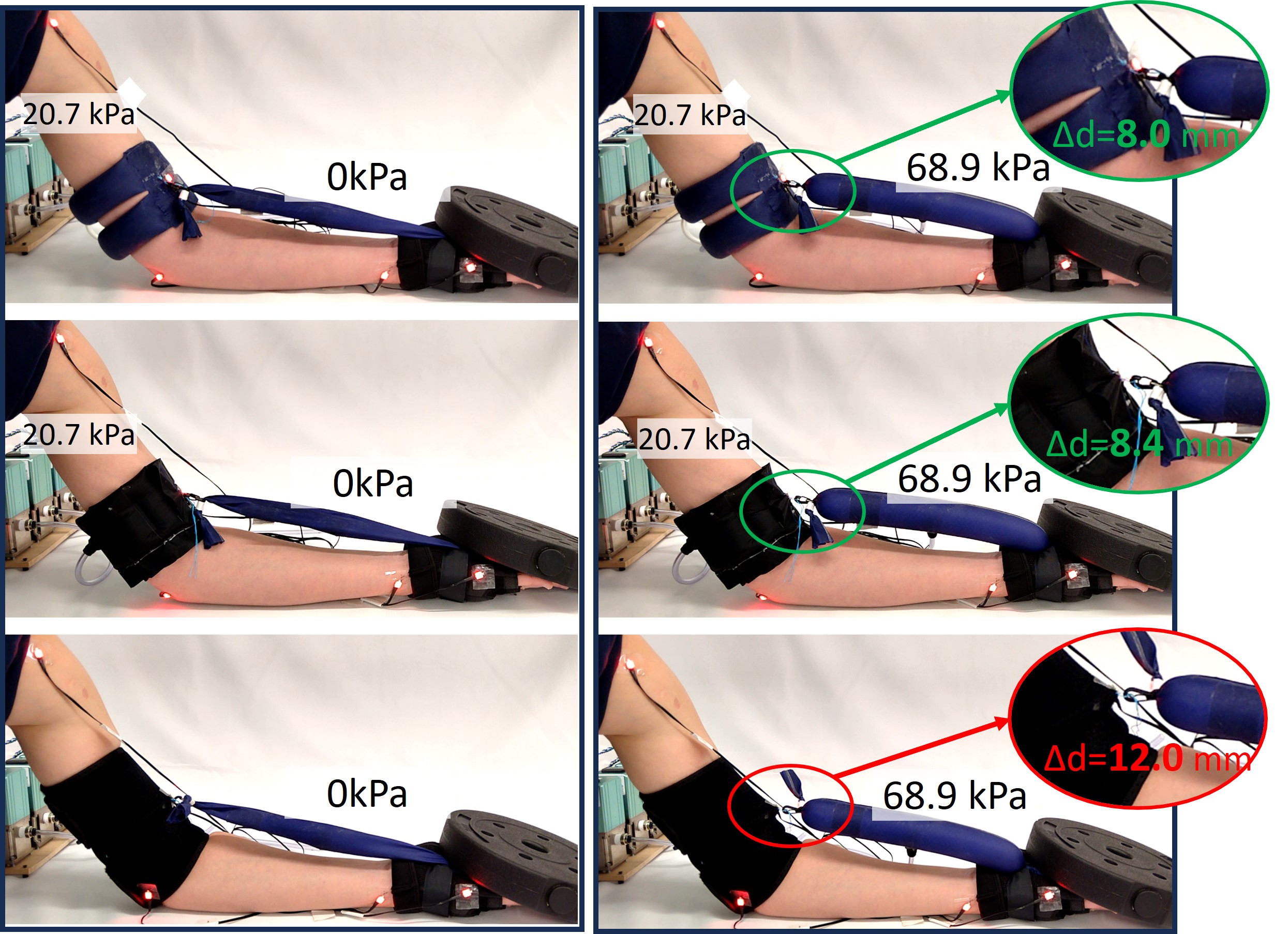}
      \caption{Comparison of three sleeves with fixed wrist position, measuring the exosuit actuator endpoint displacement ($\Delta d$) during actuator inflation. The fPAM sleeve (\textit{first row}), the SPM sleeve (\textit{second row}), and the hook and loop sleeve (\textit{third row}) shown in Fig.~\ref{three_sleeves} are incorporated in a linear fPAM-actuated wrist exosuit. To imitate a resistance exercise, the wrist movement is restricted by placing a 2~kg weight on the hand. The first column shows the sleeves anchored and the exosuit actuator uninflated. The second column shows the deformation of the sleeves when the exosuit actuator applies a pulling force; the red circle highlights the observed undesired displacement of the hook and loop sleeve, compared to the smaller displacements of the pneumatic sleeves.}
      \label{demonstration1}
      \vspace{-0.5 cm}
   \end{figure}

\edits{First, we imitated a resistance exercise with the exosuit as shown in Fig.~\ref{demonstration1} by placing a weight on the hand to fix the wrist extended. In this scenario, the linear fPAM actuator applies high pulling force on the mounting points of the actuator. The pneumatic sleeves were inflated to the highest pressure the user found comfortable (20.7~kPa), and the hook and loop band was fastened around the arm at the elbow. The actuator pressure was initially 0~kPa (first column in Fig.~\ref{demonstration1}), and then it was increased to 68.9 kPa (second column in Fig.~\ref{demonstration1}), with one trial for each sleeve. The motion capture data confirmed that the time-average of the recorded elbow and wrist joint angles was similar for each sleeve with 1.1$^\circ$ standard deviation across the sleeves for the elbow and 5.5$^\circ$ for the wrist.}

\edits{The measured mounting point positions showed that the displacement was low for the fPAM sleeve (8.0 mm) and SPM sleeve (8.4 mm) while it was significantly higher for the hook and loop sleeve (12.0 mm) due to the stretching of its material.}

\subsection{Comparison of Pneumatic Sleeves with Moving Wrist}

\begin{figure}[tb]
      \centering
      \includegraphics[width=\columnwidth]{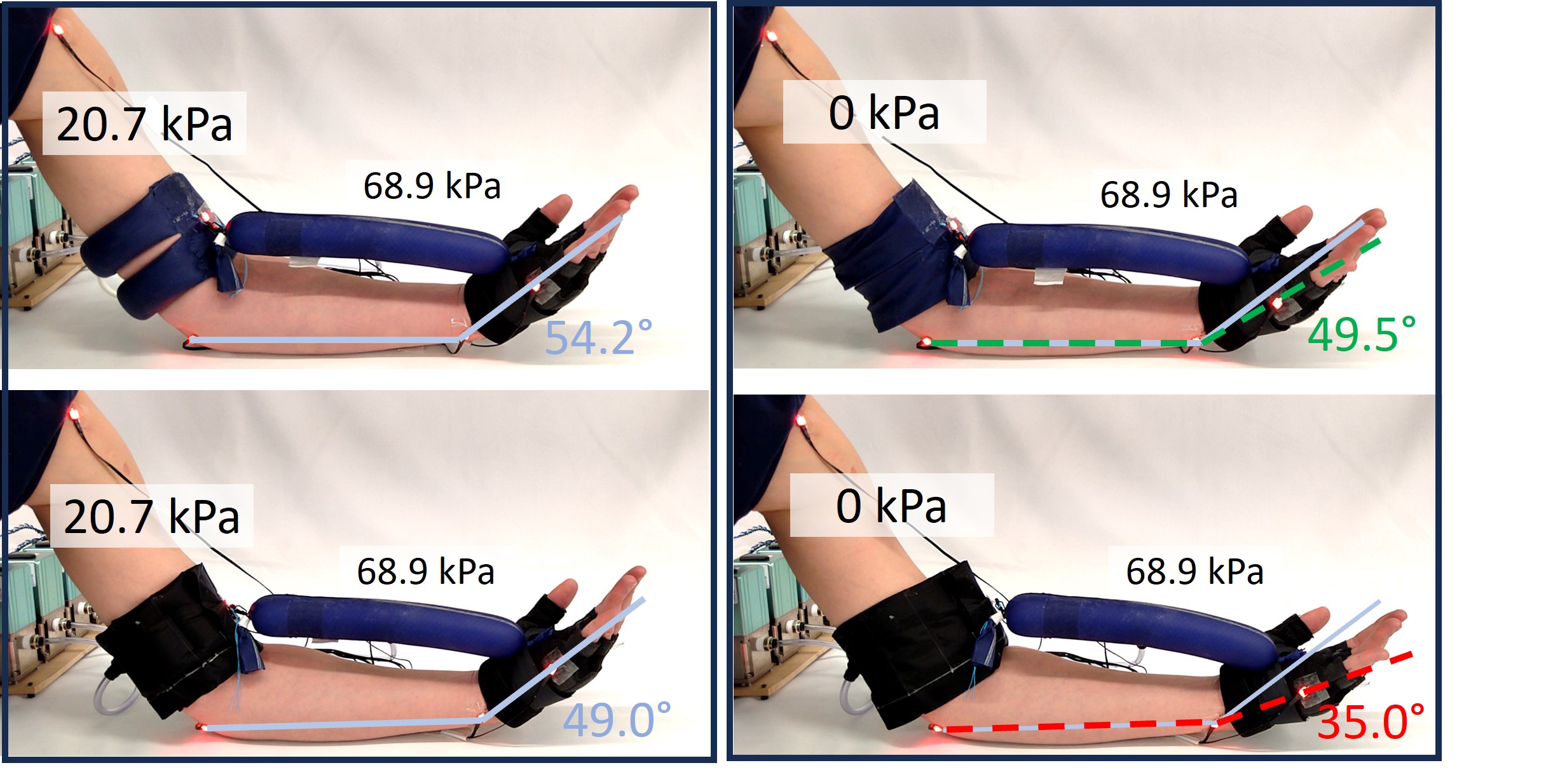}
      \caption{Comparison of pneumatic sleeves with moving wrist, measuring wrist angle change during sleeve deflation. The fPAM-actuated exosuit assists the wrist movement with the fPAM sleeve (\textit{first row}) and the SPM sleeve (\textit{second row}). The first column shows the bending of the wrist (wrist angle is denoted by light blue line) when the sleeves are inflated. The second column shows the decrease in wrist angle when the sleeves are deflated while the linear fPAM actuator pressure is unchanged. The red dotted line shows that the SPM sleeve results in a more significant wrist angle decrease (which is undesirable) than the angle decrease in case of the fPAM sleeve (denoted by green dotted line).}
      \label{demonstration2}
      \vspace{-0.5 cm}
   \end{figure}

\edits{Second, we examined the performance of the pneumatic sleeves when assisting wrist flexion. The user's wrist was relaxed during the whole demonstration process. Initially, the sleeves and the fPAM actuator were inflated and the exosuit passively bent the wrist (first column of Fig.~\ref{demonstration2}). Then, the pressure of the pneumatic bands was set to zero while the actuator pressure was unchanged, with one trial per sleeve. The resulting exosuit configuration is shown in the second column of Fig.~\ref{demonstration2}. The joint angles were computed using the motion capture data in all four cases. The elbow angles were similar across the measurements with a standard deviation across the four configurations of 2.4$^\circ$.}

\edits{When inflated, both sleeves provide firm anchoring, however, when deflated, the mounting point on the SPM sleeve undergoes high displacement, which leads to a 14.0$^\circ$ decrease in wrist angle, while the fPAM sleeve still provides low-displacement anchoring and only allows a 4.7$^\circ$ angle decrease.}

\section{Conclusions and Future Work}

\edits{In this work we explored how an fPAM-based sleeve for exosuit anchoring behaves and compares in performance to an SPM-based and an off the shelf hook and loop anchoring method. When examining various fPAM bands, measurements showed that the increase of the band width and the decrease of the band length increases both the compressing and the holding force.} 

\edits{Compared to the hook and loop sleeve, the pneumatic sleeves showed adaptability, such that their compressing force and stiffness increase with increased pressure; they can also reach higher force values. Also, we experienced that the donning/doffing process is easier for the pneumatic sleeves, especially for the SPM sleeve. Other mechanisms, such as the one presented in~\cite{Choi2019_active_anchor}, can also provide adaptability, however, the pneumatic bands have the benefits of simple design, low-cost, and easy integration into an exosuit using fluidic actuation.}

\edits{Although the resting equivalent fPAM sleeve had significantly reduced anchoring performance compared to the SPM sleeve, the fully stretched equivalent fPAM sleeve had similar behavior. The fPAM sleeve has benefits over the SPM sleeve such as its ability to exert compressing and holding force when uninflated and its increased durability, as it can hold higher pressure.}

In future work, we will focus on the modeling of the fPAM band to be able to predict the applied compressing force for a given internal pressure. To better understand the current measurement results, we are interested in exploring the effect of the localized reinforcement at the mounting points and the surface material properties of the sleeves on the mounting point displacement. Furthermore, we aim to investigate the interaction of the sleeve with the human body to ensure the user's comfort \edits{and further examine the performance of the sleeves considering these limitations that ensuring safety and comfort implies.}


\section*{Acknowledgement}

We thank Ciera McFarland for useful discussions about fabrication of series pouch motors, and we thank Tongjia Zheng for sharing resources for the motion capture data processing.

\printbibliography

\end{document}